\documentclass[10pt,twocolumn,letterpaper]{article}

\usepackage[pagenumbers]{cvpr} % To force page numbers, e.g. for an arXiv version

\usepackage{graphicx}
\usepackage{amsmath}
\usepackage{amssymb}
\usepackage{booktabs}
\usepackage{mathtools}
\usepackage{cite}
\usepackage{adjustbox}
\usepackage{multicol}
\usepackage{multirow}
\usepackage{pifont}
\usepackage{dsfont}
\usepackage{array}
\usepackage{dblfloatfix}
\usepackage{enumitem}
\usepackage{times}
\usepackage{epsfig}
\usepackage[pagebackref,breaklinks,colorlinks]{hyperref}

\usepackage[capitalize]{cleveref}
\crefname{section}{Sec.}{Secs.}
\Crefname{section}{Section}{Sections}
\Crefname{table}{Table}{Tables}
\crefname{table}{Tab.}{Tabs.}

\def\cvprPaperID{8594} % *** Enter the CVPR Paper ID here
\def\confName{CVPR}
\def\confYear{2022}
\newcommand*{\affaddr}[1]{#1}
\newcommand*{\affmark}[1][*]{\textsuperscript{#1}}
\newcommand*{\email}[1]{\tt\small{#1}}

\begin{document}

\title{The Second Place Solution for The 4th Large-scale Video Object Segmentation Challenge——Track 3: Referring Video Object Segmentation}

\author{Leilei Cao\affmark[1,2], Zhuang Li\affmark[1,3], Bo Yan\affmark[1], Feng Zhang\affmark[1],Fengliang Qi\affmark[1],Yuchen Hu\affmark[1], Hongbin Wang\affmark[1] \\
\small
\affaddr{\affmark[1]Ant Group} \quad 
\affaddr{\affmark[2]Northwestern Polytechnical University} \quad
\affaddr{\affmark[3]Tongji University} \quad
\\
\email{mcaoleilei@sina.com, jiangzi.lz@antgroup.com, hongbin.whb@antgroup.com} \\
%\and
}
\maketitle

%%%%%%%%% ABSTRACT
\begin{abstract}
The referring video object segmentation task (RVOS) aims to segment object instances in a given video referred by a language expression in all video frames. Due to the requirement of understanding cross-modal semantics within individual instances, this task is more challenging than the traditional semi-supervised video object segmentation where the ground truth object masks in the first frame are given. With the great achievement of Transformer in object detection and object segmentation,  RVOS has been made remarkable progress where ReferFormer achieved the state-of-the-art performance. In this work, based on the strong baseline framework——ReferFormer, we propose several tricks to boost further, including cyclical learning rates, semi-supervised approach, and test-time augmentation inference. The improved ReferFormer ranks 2nd place on CVPR2022 Referring Youtube-VOS Challenge.
\end{abstract}

%%%%%%%%% BODY TEXT
\section{Introduction}
Referring Video Object Segmentation (RVOS)\cite{URVOS}, a fundamental task in computer vision, aims to segment object instances in a given video referred by a language expression in all video frames. A wide range of video-related applications refer to RVOS, e.g., video editing, video surveillance, and human-object interactions. Comparing with referring image segmentation in which objects are referred to by their appearance, the objects in RVOS are referred to by the actions they are performing\cite{wu2022referformer,botach2022}. This makes solving RVOS more complicated. RVOS is also more challenging than the traditional semi-supervised video object segmentation in which the ground truth object masks in the first frame are given\cite{yang2021aot}, because it requires understanding cross-modal semantics within individual instances. In addition, objects appearance variation, occlusion, and complicated background also greatly challenge RVOS.

The traditional methods for RVOS can be divided into two categories: (1) Bottom-up methods. These methods fuse features extracted from videos and language, and then adopt a decoder\cite{long2015} to produce object masks. (2). Top-down methods. These methods first use an instance segmentation model to generate all object masks in videos to form tracklet candidates, and then use the language as the grounding criterion to select the best-matched one \cite{wu2022referformer}. With the advancement of Transformers \cite{transformer2017, dosovitskiy2021an,Liu_2021_ICCV} in object detection \cite{carion2020end,zhu2021deformable} and segmentation\cite{zheng2021rethinking,xie2021segformer,cheng2021maskformer,wang2021end,wang2021max,hwang2021video}, RVOS has been made remarkable progress \cite{wu2022referformer,botach2022,liang2022}. Botach et al.\cite{botach2022} proposed a multimodal Transformer model to process video and text together, which is end-to-end trainable and requires no additional mask-refinement post-processing steps. Wu et al. \cite{wu2022referformer} proposed a unified framework termed ReferFormer to segment and track the referred object in all frames in an end-to-end manner, in which the language expression is viewed as queries and directly attend to the most relevant regions in the video frames. And ReferFormer achieved state-of-the-art results on Ref-Youtube-VOS dataset.

In this work, based on the strong framework of ReferFormer, we propose several tricks to improve ReferFormer on RVOS. Instead of monotonically decreasing the learning rate during training, we use the cyclical learning rates to finetune the trained model of ReferFormer on the Ref-Youtube-VOS dataset. This finetuned model has performed better than the baseline model, thus the predicted results on the validation set of Ref-Youtube-VOS dataset can be served as pseudo ground truth object masks of validation set. We then re-finetune the baseline model on the training set and validation set with pseudo labels. This semi-supervised approach is also employed on the testing set. The improved ReferFormer ranks 2nd place in the 4th Large-scale Video Object Segmentation Challenge (CVPR2022)——Track 3: Referring Video Object Segmentation, with an overall $\mathcal{J}\&\mathcal{F}$ of 68.6\% and 61.7\% on \texttt{test}-\texttt{dev} and \texttt{test}-\texttt{challenge}, respectively.
%-----------------------------------------------------------------
\begin{figure}
  \centering
  \includegraphics[width=1.0\linewidth]{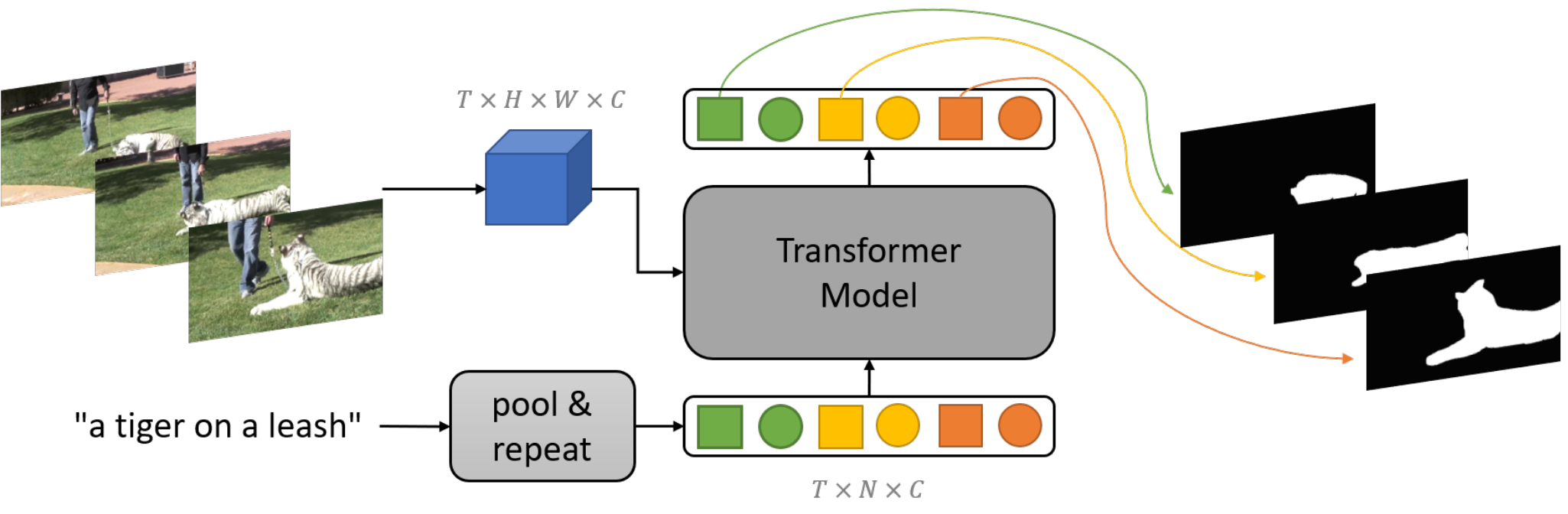}
  \caption{The overall pipeline of ReferFormer\cite{wu2022referformer}, where the language expression is served as conditional queries to focus on the referred object. The detailed architecture of ReferFormer can refer to \cite{wu2022referformer}.}
  \label{fig:pipeline}
\end{figure}

\section{Method}
Our model is finetuned based on ReferFormer, and the overall pipeline of ReferFormer is illustrated in Figure \ref{fig:pipeline}. Unlike the traditional methods, ReferFormer treats the language expression as conditional queries to focus on the referred object. And the queries are viewed as instance-aware dynamic kernels to filter out the segmentation masks. The details of ReferFormer can refer to \cite{wu2022referformer}.
To improve the performance of ReferFormer on Ref-Youtube-VOS dataset, we propose several tricks to finetune the model, including cyclical learning rates, semi-supervised approach and test time augmentation (TTA) inference.

\begin{figure}
  \centering
  \includegraphics[width=1.0\linewidth]{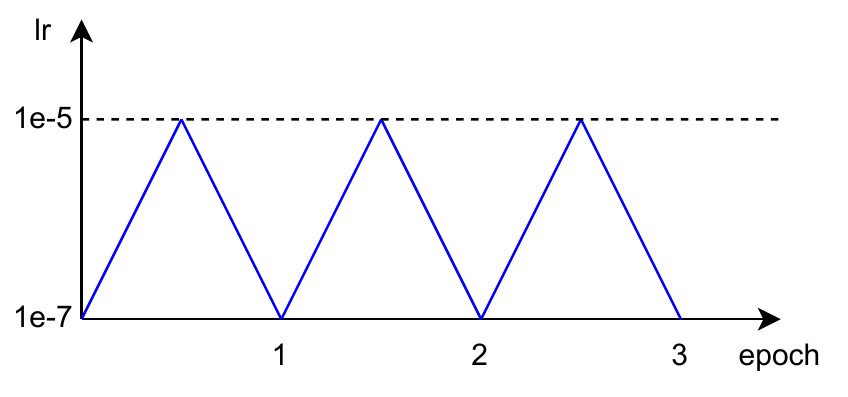}
  \caption{Cyclical learning rates.}
  \label{fig:lr}
\end{figure}

\subsection{Cyclical Learning Rates}
In the original paper of ReferFormer, the model was first pretrained on the image referring segmentation datasets Ref-COCO\cite{yu2016}, Ref-COCOg\cite{yu2016} and Ref-COCO+\cite{mao2016}. Then, the model was finetuned on Ref-Youtube-VOS dataset. In the latest version of their released code, the model was joint trained with the image referring segmentations datasets, which achieved better results than pretraining. Since of strong performance of ReferFormer, we directly finetune the trained model on Ref-Youtube-VOS dataset. Instead of using fixed values of learning rates or monotonically decreasing the learning rates during finetuning, we use cyclical learning rates (CLR)\cite{smith2015} as illustrated in Figure \ref{fig:lr}. We use the triangular learning rate policy, in which the learning rate linearly increases to the maximum boundary and decreases to the minimum boundary. The length of a cycle is set to one epoch, and the minimum and maximum learning rate is set to 1e-7 and 1e-5, respectively. 

\subsection{Semi-Supervised Approach}
We use a semi-supervised training method to finetune the model\cite{ainnoseg}. There are several steps to follow. 
\begin{itemize}[leftmargin=*]
\vspace{-2mm}
\item Step 1: We use CLR to finetune the trained model of ReferFormer on Ref-Youtube-VOS dataset. 
\vspace{-2mm}
\item Step 2: The finetuned model predicts object masks of validation-set, thus forming a validation-set composed of all videos frames and pseudo ground truth pairs. 
\vspace{-2mm}
\item Step 3: We re-finetune the model on training-set and joint with validation-set with pseudo ground truth. 
\vspace{-2mm}
\item Step 4: We re-predict the validation-set again to acquire better pseudo ground truth. The new pseudo ground truth and frames pairs can be used to complete a second-round re-finetuning. 
\vspace{-2mm}
\item Step 5: The two-round finetuned model predicts testing-set to form pseudo ground truth labels and videos frames pairs. The validation-set and testing-set with pseudo ground truth labels and training-set are joint used to re-finetune the model.
\vspace{-2mm}
\item Step 6: Finally, the model predicts testing-set to get better results. 
\vspace{-2mm}
\end{itemize}

%--------------------------------------------------------------

\section{Experiments}
\subsection{Training Details}
We follow the training details of original ReferFormer \footnote{\textcolor{magenta}{https://github.com/wjn922/ReferFormer}}, except being specified. The baseline trained model of ReferFormer that we selected is the model with a Video-Swin-B backbone\cite{Liu_2021_ICCV,liu2021video} joint trained with Ref-COCO/+/g datasets, this model achieves the overall $\mathcal{J}\&\mathcal{F}$ of 64.9\% on the validation-set. We finetune and evaluate our model with 8 A100 GPUs on Ref-Youtube-VOS dataset.

\begin{table*}[ht]
\centering
\caption{Experimental results of baseline ReferFormer on validation-set using test time augmentation inference.}
\label{tab:tta}
%\vspace{4pt}
\scalebox{0.93}{
\begin{tabular}{l|c|c} 
\hline
model  & size of frames & $\mathcal{J}\&\mathcal{F}$  \\
\hline
baseline & 360 & 64.9   \\
\hline
+ horizontal flip & 360 & 65.5 \\
\hline
+ multi-scale inference & 288,352,448,512,640 & 65.8 \\
\hline
+horizontal flip + multi-scale inference & 288,352,448,512,640 & 66.6 \\
\hline
\end{tabular}
}
\vspace{-2mm}
\end{table*}

\begin{table}[ht]
\centering
\caption{Experimental results of ReferFormer finetuned with cyclical learning rates. }
\label{tab:clr}
%\vspace{4pt}
\scalebox{0.93}{
\begin{tabular}{l|c|c} 
\hline
model  & size of frames & $\mathcal{J}\&\mathcal{F}$  \\
\hline
baseline & 360 & 64.9   \\
\hline
+ CLR & 360 & 65.9 \\
\hline
+CLR+TTA & 288,352,448,512,640 & 67.2 \\
\hline
\end{tabular}
}
\vspace{-2mm}
\end{table}

\begin{table}[ht]
\centering
\caption{Experimental results on validation-set of ReferFormer finetuned with semi-supervised approach on the first-round finetuning.}
\label{tab:ss_f}
%\vspace{4pt}
\scalebox{0.93}{
\begin{tabular}{l|c|c} 
\hline
model  & size of frames & $\mathcal{J}\&\mathcal{F}$  \\
\hline
baseline & 360 & 64.9   \\
\hline
+ semi-supervised & 360 & 66.8 \\
\hline
+semi-supervised+TTA & 288,352,448,512,640 & 68.0 \\
\hline
\end{tabular}
}
\vspace{-2mm}
\end{table}

\begin{table}[ht]
\centering
\caption{Experimental results on validation-set of ReferFormer finetuned with semi-supervised approach on the second-round finetuning.}
\label{tab:ss_s}
%\vspace{4pt}
\scalebox{0.93}{
\begin{tabular}{l|c|c} 
\hline
model  & size of frames & $\mathcal{J}\&\mathcal{F}$  \\
\hline
baseline & 360 & 64.9   \\
\hline
+ semi-supervised +flip & 720 & 68.3 \\
\hline
+semi-supervised+TTA & 352,512,720,896 & 68.6 \\
\hline
\end{tabular}
}
\vspace{-2mm}
\end{table}

\begin{table}[ht]
\centering
\caption{Experimental results on testing-set of ReferFormer finetuned with semi-supervised approach.}
\label{tab:ss_t}
%\vspace{4pt}
\scalebox{0.93}{
\begin{tabular}{l|c|c} 
\hline
model  & size of frames & $\mathcal{J}\&\mathcal{F}$  \\
\hline
w/o semi-supervised + flip & 720 & 61.2   \\
\hline
+ semi-supervised + flip & 720 & 61.7 \\
\hline
\end{tabular}
}
\vspace{-2mm}
\end{table}
 
\begin{table}[ht]
\centering
\caption{Testing results on CVPR2022 Referring-YouTube-VOS Challenge.}
\label{tab:challenge}
%\vspace{4pt}
\scalebox{0.93}{
\begin{tabular}{l|ccc} 
\hline
Team  & $\mathcal{J}\&\mathcal{F}$ & $\mathcal{J}$  & $\mathcal{F}$  \\
\hline
Bo$\_\_\_\_$ & 64.1 & 62.2 & 66.1  \\
\hline
\textbf{jiliushi (Ours)} & 61.7 & 59.8 & 63.6 \\
\hline
PENG & 60.8 & 58.9 & 62.7 \\
\hline
ds-hohhot & 59.6 & 57.9 & 61.2 \\
\hline
JQK &  59.4 & 57.7 & 61.1 \\
\hline
nero & 58.0 & 56.1 & 59.9 \\
\hline
\end{tabular}
}
\vspace{-2mm}
\end{table} 

\subsection{Components Analysis}
\paragraph{Test Time Augmentation Inference.}
We evaluate validation-set based on the baseline trained model using test time augmentation inference, as shown in Table \ref{tab:tta}. We first evaluate using single-scale (the short side size of frames is set as 360) inference with horizontal flip, which boosts 0.6pt comparing with the baseline. Then we evaluate using multi-scale inference (the short side size of frames is set as 288,352,448,512,640), which boosts 0.9pt. Finally, we evaluate using multi-scale inference with horizontal flip to achieve the overall $\mathcal{J}\&\mathcal{F}$ of 66.6\%.
\paragraph{Cyclical Learning Rates.}
As shown in Table \ref{tab:clr}, we finetune the baseline model using CLR for 4 epochs to achieve the best performance, which brings 1.0pt improvement using single-scale inference and 2.3pt improvement using TTA inference, respectively. 
\paragraph{Semi-supervised Approach.}
We first predict object masks of validation-set as pseudo ground truth using the model which achieving the overall score of 67.2\%. Then we re-finetune the baseline model on training-set and joint with validation-set with pseudo ground truth for 5 epochs using CLR. After that, we evaluate the re-finetuned model as shown in Table \ref{tab:ss_f}, which boosts 1.9pt with single-scale inference and 3.1pt with TTA comparing with the baseline, respectively. We then re-predict object masks of validation-set as new pseudo ground truth, and re-finetune again. In this second-round re-finetuning, all frames are downsampled so that the short side has the size of 608, 672 and 720 and the maximum size for long side is 1280. We re-finetune for 7 epochs using CLR and evaluate again as shown in Table  \ref{tab:ss_s}. The performance of ReferFormer has been improved to 68.3\% with single-scale (720) inference and horizontal flip, and it is been further improved to 68.6\% with TTA inference. The second-round finetuned model can directly predict testing-set with single-scale (720) inference and horizontal flip and get the overall score of 61.2\% as shown in Table \ref{tab:ss_t}. To improve the model on testing-set, the predicted object masks of testing-set can also be served as pseudo ground truth, and we re-finetune the baseline model using CLR for 7 epochs on training-set joint with validation-set and testing-set. Finally, we evaluate the last model on testing-set with single-scale (720) inference and horizontal flip, which achieves the overall $\mathcal{J}\&\mathcal{F}$ of 61.7\%.
\subsection{Challenge Results}
We submit our results to CVPR2022 Referring Youtube-VOS Challenge, which ranks second place as shown in Table \ref{tab:challenge}. It is worth noting that we did not use model ensemble and multi-scale inference since of time limited. And we only implement one-round re-finetuning on testing-set with pseudo ground truth. Using these tricks, the performance of RerferFormer may be improved further.
%--------------------------------------------------------------
\section{Conclusion}
In this paper, we proposed several tricks to improve ReferFormer performing on Refer-YouTube-VOS dataset, including cyclical learning rate, semi-supervised approach, and test time augmentation inference. The boosted model achieves an overall $\mathcal{J}\&\mathcal{F}$ of 61.7\% on testing-set, ranking second place on CVPR2022 Referring Youtube-VOS Challenge.

%%%%%%%%% REFERENCES
{\small
\bibliographystyle{ieee_fullname}
\bibliography{egbib}
}

\end{document}